# Applications of Machine Learning in Detecting Afghan Fake Banknotes

Hamida Ashna, Ziaullah Momand
Faculty of Computer
Department of Information Systems
Kabul University
Kabul,1001, Afghanistan
hamidaashna123@gmail.com, mommand.csf@gmail.com

*Abstract*— Fake currency, unauthorized imitation money lacking government approval, constitutes a form of fraud. Particularly in Afghanistan, the prevalence of fake currency poses significant challenges and detrimentally impacts the economy. While banks and commercial establishments employ authentication machines, the public lacks access to such systems, necessitating a program that can detect counterfeit banknotes accessible to all. This paper introduces a method using image processing to identify counterfeit Afghan banknotes by analyzing specific security features. Extracting first and second order statistical features from input images, the WEKA machine learning tool was employed to construct models and perform classification with Random Forest, PART, and Naïve Bayes algorithms. The Random Forest algorithm achieved exceptional accuracy of 99% in detecting fake Afghan banknotes, indicating the efficacy of the proposed method as a solution for identifying counterfeit currency.

*Keywords—Fake Afghan banknotes, First order features, MATLAB, Naïve Bayes, PART, Random Forest, second order features, WEKA.*

## I. Introduction

Currency is a universally accepted medium of exchange that enables the trade of goods and services. Typically, in the form of paper and coins, it represents a government-issued monetary system and holds a face value that serves as a means of payment. The establishment of currency has greatly facilitated global trade and has been in existence for over 3,000 years. While its origins can be traced back to Ancient Egypt, the evolution of currency has led to its current form and widespread usage [1, 2].

Currently, there are more than 200 national currencies in circulation worldwide. Remarkably, 42 countries either utilize the U.S. dollar or have their currencies directly pegged to it. The U.S. dollar holds a significant position in the global economy, accounting for 58.8% of foreign exchange reserves, according to the International Monetary Fund (IMF). Most nations have their own official currencies, such as the Swiss franc for Switzerland and the yen for Japan. However, the Euro stands as an exception, having been adopted by the majority of European Union member states [1]. In terms of value, the top five highest-valued currencies globally are the Kuwaiti Dinar, Bahraini Dinar, Omani Rial, Jordanian Dinar, and Cayman Islands Dollar [3].

Afghanistan, like any other country, possesses its official currency to facilitate domestic transactions. In contrast to widely used currencies like the Euro (EUR) or Dollar (USD), Afghanistan has its distinct national currency called the Afghani (AFN). The issuance of the Afghani is overseen by the country's central bank, Da Afghanistan Bank, and its adoption and popularity have been shaped by a significant historical background. The Afghani is available in the form of bills and coins. Bill denominations include 10, 20, 50, 100, 500, and 1,000 AFN, while coin denominations include 1, 2, and 5 AFN [4].

The usage of currency has brought numerous conveniences to society. However, along with its benefits, the challenge of identifying genuine currency arises due to the existence of counterfeit or fake money. This problem extends beyond Afghanistan and is faced by countries worldwide, as counterfeit denominations pose a significant issue in currency recognition. Counterfeiting has a long-standing history and is often referred to as the world's second-oldest profession. Counterfeit money refers to currency created without the legal authorization of a government or state, typically with the intention of imitating genuine currency and deceiving its recipients. The production or usage of counterfeit money is considered a form of forgery and is strictly prohibited by law [5].

The presence of counterfeit currency circulating in an economy has a significant impact on the stability and value of legitimate currency, posing a threat to its overall integrity. With advancing technology, the risk of increased production and dissemination of counterfeit banknotes grows, which can have detrimental effects on a nation's economy. Therefore, it is imperative to develop methods to detect counterfeit currency. Afghanistan, in particular, faces a high prevalence of counterfeit currency, causing numerous challenges for its people. While banks and commercial establishments in Afghanistan employ authentication machines, these systems are not accessible to the public. To address this issue, we propose a method to identify counterfeit Afghan banknotes by examining specific security features using image processing and machine learning approach.

As previously mentioned, Afghani banknotes are available in six different denominations. However, it has been observed that counterfeit banknotes predominantly occur in the 500 and 1000 AFN denominations [6]. In the market, various banknotes have been issued and are in circulation in Afghanistan, including those issued in 1387, 1389, 1391, and 1395. In this study, our focus was specifically on the 1000 AFN banknotes issued in 1391.

The remaining sections of this paper are organized as follows: In Section II, we discuss the related work pertaining to our study. The methodology employed in our work is described in Section III. Section IV presents the results obtained from our analysis, and in Section V, a comprehensive discussion of these results is provided. Concluding remarks are presented in Section VI, along with an overview of our future research directions.

## II. RELATED WORK

In recent years, notable advancements have been achieved in the field of counterfeit currency detection, leading to the implementation of dedicated detection systems and software by various countries worldwide. Notable examples include India, Libya, Indonesia, and Egypt, among others [7, 8, 11, 13]. These countries have developed their own robust systems to effectively differentiate between genuine and counterfeit currencies, showcasing the global efforts in tackling this issue.

The identification of counterfeit currency poses unique challenges for researchers due to the distinct characteristics of coins and banknotes. A comprehensive review of recent literature reveals the existence of various successful approaches to detect fake banknotes. These methods encompass a range of techniques, including deep learning algorithms, image processing, machine learning, and hybrid combinations of these approaches. The diversity of these methods highlights the ongoing efforts to develop effective strategies for counterfeit currency detection.

In a notable study, the authors devised an affordable and efficient system for the identification of Indian banknotes. The system utilized image processing techniques and focused on extracting six key features from the banknote image: an identification mark, security thread, watermark, numeral, floral design, and micro lettering [7]. Likewise, in another research endeavor, a system was proposed to verify the authenticity of Egypt banknotes, leveraging the capabilities of the MATLAB platform. The proposed approach in this study employed two feature vectors extracted from the banknote image: texture and shape. Texture features were extracted using the Gray-Level Co-occurrence Matrix (GLCM), while shape features were obtained using a set of common properties that characterize connected image regions [8]. This combination of texture and shape analysis enables a comprehensive evaluation of the banknote image, enhancing the accuracy and effectiveness of the proposed approach. In another study, a fake currency detection system was developed using MATLAB's image processing capabilities. The system effectively detected counterfeit currency in the newly introduced denominations of 500 and 2000, employing a comprehensive process from image acquisition to feature intensity calculation [9]. In a distinct study, researchers presented a system that verifies the genuineness of banknotes by analyzing color, texture, shape, and other distinct characteristics, as per the guidelines set by the Reserve Bank of India (RBI). The system exhibited remarkable enhancements in efficiency, achieving an average accuracy rate of approximately 89% [10].

In a related study an algorithm was developed using image processing techniques to detect counterfeit Libyan banknotes. The algorithm utilized Hu moments and comparison parameters to distinguish between genuine and fake notes [11]. Another study introduced a currency detection system for Indian banknotes, achieving a high accuracy rate of 90% using digital image processing techniques with OpenCV [12].

The authors conducted a study to enhance the authenticity of the Rupiah currency by designing a system. To identify genuine currency, they employed the K-Nearest Neighbors algorithm, while for texture feature extraction, they utilized GLCM. In their research, they focused on six GLCM features, namely angular second moment, contrast, correlation, variance, inverse different moment, and entropy [13]. Similarly, the authors developed an android application specifically designed to detect counterfeit currency. They employed MATLAB to extract security threats and achieved improved results by utilizing SVM algorithms [14]. Furthermore, another study was conducted to detect counterfeit currency by extracting first-order and second-order statistical features from currency images. The authors successfully employed an SVM classifier to analyze the feature vectors, resulting in an impressive accuracy rate of 95.8% [15]. In a relevant study, researchers presented a novel approach for differentiating between genuine and counterfeit banknotes. They employed statistical-based features and employed edge detection methods for accurate feature extraction. The extracted features were subsequently fed into an SVM classifier, enabling the system to successfully distinguish between real and fake banknotes with high accuracy[16]. Moreover, an innovative technique was proposed specifically for detecting counterfeit Indian banknotes. The researchers utilized a region of interest (ROI) cropping method to isolate individual features. Each feature vector obtained through this process was then employed to train a machine learning model. Their dataset consisted of a comprehensive collection of real banknote images, ensuring a robust analysis [17].

Previous studies have extensively utilized image processing techniques and machine learning algorithms to detect counterfeit banknotes. However, there is a notable gap in research regarding the detection of fake Afghan banknotes. Unlike Afghanistan, numerous countries have seen advancements in automatic currency note recognition systems, dataset creation, and comprehensive literature. Consequently, the aim of this study was to address this research gap and contribute by identifying and proposing an improved machine learning algorithm specifically tailored for the detection of counterfeit Afghan banknotes.

## III. METHODOLOGY

Banknotes possess various distinguishing features, including color, texture, size, watermarks, and security measures. While certain features like color, size, and texture are readily visible, others like watermarks remain concealed. Enhancing the speed and accuracy of banknote detection models necessitates the extraction of a minimal set of features. In this proposed method, we focus on extracting only two key features from Afghan banknotes. The detection methodology for Afghan banknotes is comprised of multiple stages, each contributing to the overall process. Figure 1 provides an overview of the proposed methodology, highlighting its key components.

### A. Image Acquisition

In the proposed method, high-resolution images were acquired using a scanner with a resolution of 600 dpi, saving them in the .jpg format for subsequent analysis. Figure 2 illustrates the process of collecting images specifically from 1000 AFN banknotes that were issued in 1391, encompassing both genuine and counterfeit banknotes.

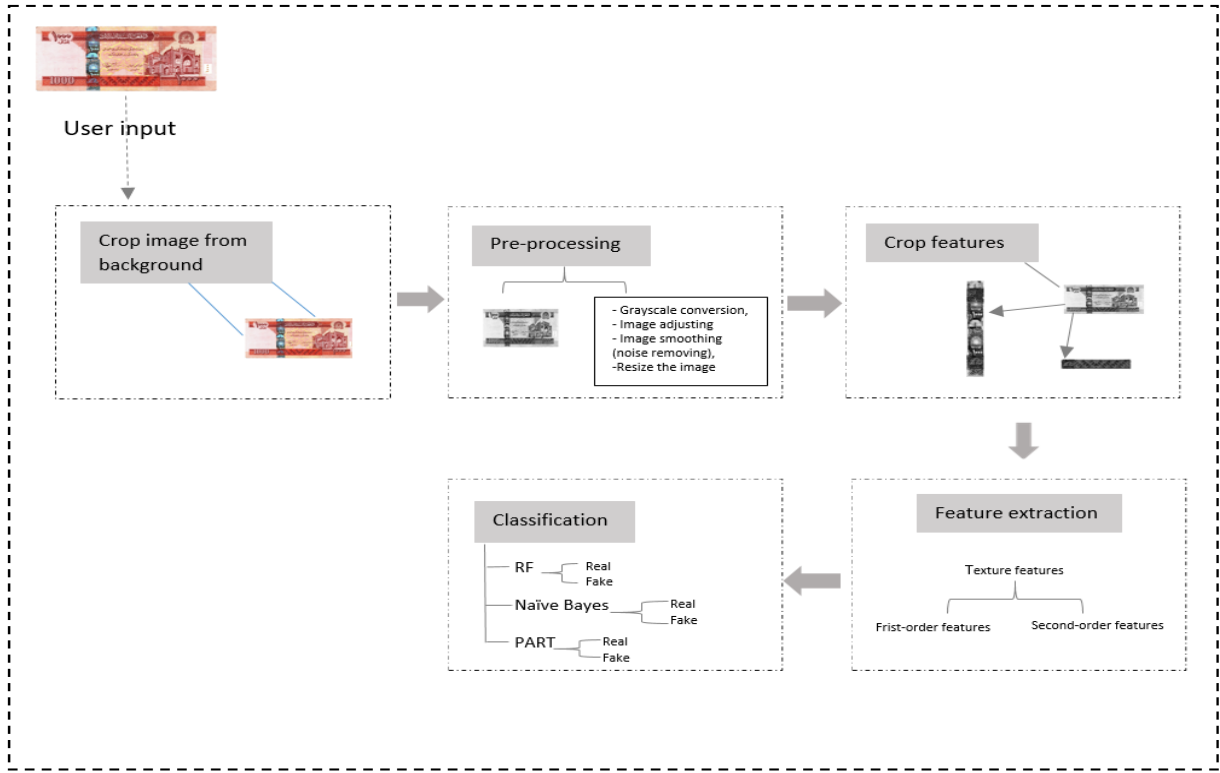

Fig. 1. Flow of proposed method

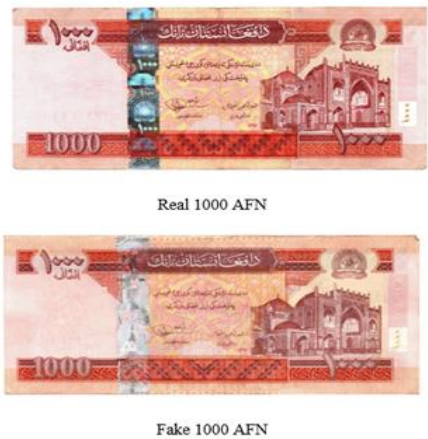

Fig. 2. Figure illustrating the Real and Fake 1000 AFN banknotes.

### B. Crop image from background

During the image acquisition process, it is common to capture irrelevant background information along with the banknote image. To ensure accurate identification and recognition of the banknote, it is crucial to remove this unwanted background. In this study, we employed a technique to eliminate the background by identifying the largest blob within the image. To achieve this, we utilized thresholding and masking techniques to isolate the largest blob in the image. The boundary coordinates of this blob were then determined, and the image was cropped using MATLAB's "imcrop" function. This cropping process effectively isolated the banknote, allowing us to focus solely on the banknote's features while eliminating any interference from the background.

### C. Pre-processing

To enhance the feature extraction efficiency, we applied pre-processing steps to clean up the banknote images. This involved grayscale conversion, image adjustment, noise removal, and resizing. These measures address discontinuities, illumination issues, noise, and varying sizes, ensuring optimal image quality for further analysis.

The captured image is initially in RGB color format, which can be computationally heavy and contain more noise. To address these challenges, the image is converted to grayscale. This conversion reduces the image size and simplifies the processing by focusing on intensity information instead of handling separate red, green, and blue components. Furthermore, removing color information improves the feature extraction process. Additionally, the original image size obtained from the scanner was typically large. To optimize processing efficiency, the image was resized to 1056x2481 pixels. Removing noise is a crucial step in image processing to prevent interference with subsequent analysis. In this study, the Wiener filter was applied to effectively eliminate noise.

### D. Crop features

The image processing stage involves cropping out the key point features from the banknote image, known as region of interest (ROI) cropping. These cropped images are considered as individual features. The extracted set of features was then fed into a machine learning algorithm for prediction. In this study, the focus was on extracting 2 main features from the Holographic strip and bottom design of the 1000 AFN banknote image, as depicted in figure 3.

### E. Feature Extraction

Banknote detection methods often utilize color, shape, and texture as distinguishing factors, with various established techniques for feature extraction. In this study, the focus was on utilizing texture statistical-based feature extraction methods. Specifically, the attention was directed towards extracting first-order histogram-based features and second-order co-occurrence matrix features from banknote images.

These two types of features were chosen due to their ability to effectively capture texture and pattern information, which can serve as indicators of banknote authenticity. The research findings revealed notable differences in texture between fake and genuine AFN banknotes. Hence, combining these two feature types offers a more comprehensive representation of the banknote image, ultimately enhancing the accuracy of fake banknote detection systems.

*1) First-Order statistics:* In this study, the histogram features, also known as first-order statistical features, were directly computed from the cropped gray-level images. By analyzing the probability density of grayscale intensity levels, various informative quantitative first-order statistical features can be derived. Specifically, we focused on four key features: kurtosis, skewness, variance, and entropy. These features were extracted for both the Holographic strip and bottom design of the banknote image. The variance provides a measure of the intensity changes around the mean, while skewness quantifies the degree of asymmetry around the mean in the histogram. Kurtosis, on the other hand, measures the degree of outliers in the histogram, and entropy assesses the randomness of the intensity distribution. These four features were selected based on their relevance in previous research [13, 16, 18], as they contribute valuable information for distinguishing between genuine and counterfeit banknotes.

*2) Second-Order statistics:* The statistical features, which used in this work, are contrast, correlation, homogeneity, and energy. Energy is a feature that measures the smoothness of the image. Homogeneity is a measure that takes high values for low-contrast images. Correlation is a measure of gray level liner dependency between the pixels at the specific position relative to each other. And contrast is a measure of contrast or local intensity [13, 16, 18]. In this work, the GLCM was calculated from the grayscale version of the images in four directions(0°, 45°, 90° and 135°) using the "graycomatrix" function. And from them the statistical parameters of the GLCM were derived. After that, the mean value of each statistical parameter of the GLCM were computed. And these features were extracted for both the Holographic strip and bottom design of the banknote image.

### F. Dataset

The dataset used in this project consisted of primary data. While we collected the real banknotes from the banks, due to government policies, we had to collect the fake banknotes ourselves. As a result, the number of fake banknotes in the dataset is limited. In total, we collected 70 banknote images, with 20 of them being fake and 50 of them being real notes. The dataset was created based on the texture features extracted from two main parts of the banknote image: the Holographic strip and bottom design. Figure 4 illustrates the extracted features during the feature extraction step. The dataset comprises 17 attributes, where the first 16 attributes define the banknote characteristics. The 17th attribute represents the label, with "yes" indicating a real banknote and "no" indicating a fake banknote. It is important to note that the first 16 attributes are numeric, while the 17th attribute is nominal. Further details regarding the dataset attributes can be found in Table I.

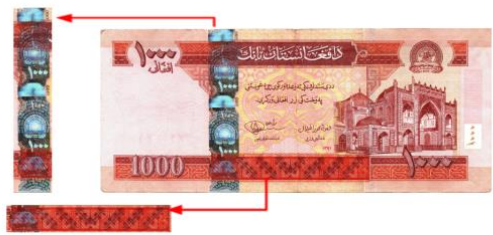

Fig. 3. AFN banknote cropped two main features

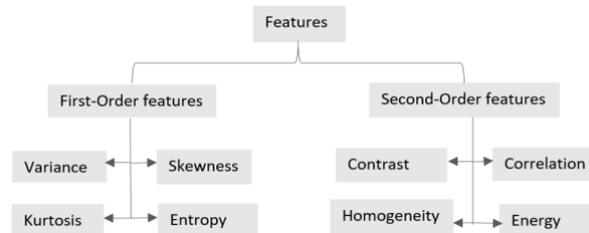

Fig. 4. Extracted texture features from AFN banknote image

TABLE I. DATASET ATTRIBUTES

| No | Features | Type | Description |
|---|---|---|---|
| 1 | Contrast_1 | Numeric | Contrast of Holographic strip |
| 2 | Correlation_1 | Numeric | Correlation of Holographic strip |
| 3 | Energy_1 | Numeric | Energy of Holographic strip |
| 4 | Homogenetiy_1 | Numeric | Homogeneity of Holographic strip |
| 5 | Entropy_1 | Numeric | Entropy of Holographic strip |
| 6 | Variance_1 | Numeric | Variance of Holographic strip |
| 7 | Skewness_1 | Numeric | Skewness of Holographic strip |
| 8 | Kurtosis_1 | Numeric | Kurtosis of Holographic strip |
| 9 | Contrast_2 | Numeric | Contrast of bottom design |
| 10 | Correlation_2 | Numeric | Correlation of bottom design |
| 11 | Energy_2 | Numeric | Energy of bottom design |
| 12 | Homogenetiy_2 | Numeric | Homogeneity of bottom design |
| 13 | Entropy_2 | Numeric | Entropy of bottom design |
| 14 | Variance_2 | Numeric | Variance of bottom design |
| 15 | Skewness_2 | Numeric | Skewness of bottom design |
| 16 | Kurtosis_2 | Numeric | Kurtosis of bottom design |
| 17 | Class | Nominal | Indicates fake and real notes |

### G. Data Processing

In this work, the WEKA machine learning tool was used for data preprocessing and modeling. It is crucial to preprocess the data before building a machine learning model in order to obtain the desired results. There are several techniques that can be utilized for data preprocessing, such as removing abnormalities and modifying characteristics. We noted the following points regarding our data: a) The dataset does not have any missing value, b) All the input features are continuous and have different ranges, so we need to scale all features in a specific way, and c) The dataset was not appropriate, and it was imbalanced.

*1) Normalization:* Normalization is usually applied to convert an image's pixel values to a typical or more familiar sense. All the input features had continuous values with different ranges. The Max-Min normalization was performed to scale all features in the range of 0 and 1 as shown in (1).

$$X_{norm} = \frac{X_{current} - X_{min}}{X_{max} - X_{min}} \qquad (1)$$

Where $X_{current}$ is the current value in the dataset, $X_{min}$ is the minimum value in the dataset, and $X_{max}$ is the maximum value in the dataset.

*2) Data Resampling:* The dataset was imbalanced due to the rarity of fake banknotes, the number of fake notes were less than real notes, which can leas to a biased model that was unable to accurately classify banknotes. The class imbalance ratio was 0.4. To address this issue, we need to balance our dataset by generating synthetical samples.

The Synthetic Minority Oversampling technique (SMOTE) was utilized. SMOTE is employed to prevent overfitting, which can occur when exact replicas of minority instances are added to the main dataset. Instead, SMOTE takes a subset of data from the minority class and generates new synthetic instances that resemble the existing ones. These synthetic instances are then added to the original dataset, creating a balanced training set for classification models. This technique ensures that no valuable information is lost in the process. Based on these considerations, SMOTE was chosen for its effectiveness in addressing class imbalance in this work [20].

The usage of SMOTE depends on the imbalance between classes in the data. In this experiment, SMOTE was employed three times, as depicted in Figure 5, ranging from 100% to 300%, until satisfactory results were achieved.In our dataset, we initially collected 20 fake banknotes and 50 real banknotes. Applying SMOTE at 100% resulted in synthetically resampling the records to 40 fake notes and 50 real notes. When SMOTE was applied at 200%, the records were increased to 120 fake notes and 99 real notes. Finally, with SMOTE at 300%, the dataset consisted of 180 fake notes and 200 real notes.

*H. Classification*

In this work, the classification model was used to classify the fake and real Afghan banknotes. Our task was a binary classification task that categorizes new observations into one of two classes (real notes or fake notes). In this study, 3 machine learning algorithms namely, Random Forest (RF), Naïve Bayes (NB), and PART were incorporated to classify the real and fake Afghan banknotes. It is notable, that PART classifier is a rule learning classifier combines the divide-and-conquer strategy with separate-and-conquers strategy. It builds a partial decision tree on the current set of instances and creates a rule from the decision tree [21]. 10-fold cross validation was applied to evaluate the model performance.

## IV. RESULTS

As the dataset was imbalance, the satisfied results for RF, NB and PART classifiers were obtained after applying SMOTE up to 300% as shown in table II. Besides accuracy, the models were evaluated based on TP Rate, Precision, Recall and F-Measure evaluation measurements. Table III, IV, and V show the results of the mentioned measurement for RF, NB, and PART classifiers, respectively. As shown in Table II, all classifiers performed well, but RF classifier obtained the outstanding results with 99% accuracy. NB classifier also obtained high accuracy of 97%, while PART achieved 96.70%. RF also performed with outstanding result for TP Rate of 99%, precision of 99%, recall of 99%, and f-measure with 99% as shown in table III.

The NB classifier boasts impressive performance regarding TP rate, precision, recall, and f-measure, as indicated in Table IV, where it achieved scores of 97.30%, 97%, 98%, and 98%, respectively. Similarly, the PART classifier, as shown in Table V, achieved scores of 97%, 97.2%, 97%, and 96.5%, respectively, for TP rate, precision, recall, and F-measure.

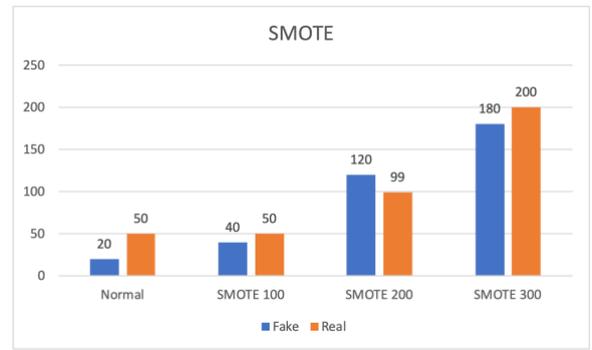

Fig. 5. Data is resampled with SMOTE in 3 iterations.

TABLE II. ACCURACY RESULTS OF ALL CLASSIFIERS WITH SMOTE

| No | SMOTE | RF | NB | PART |
|---|---|---|---|---|
| 1 | 100% | 96.43% | 92.86% | 96.43% |
| 2 | 200% | 98.08% | 96.15% | 98.08% |
| 3 | 300% | 99.00% | 97.00% | 96.70% |

TABLE III. RF CLASSIFIER RESULTS WITH SMOTE

| SMOTE | TP Rate | Precision | Recall | F-Measure |
|---|---|---|---|---|
| 100% | 96.4% | 96.6% | 96.4% | 96.4% |
| 200% | 98.1% | 98.2% | 98.1% | 98.1% |
| 300% | 99.0% | 99.0% | 99.0% | 99.0% |

TABLE IV. NAÏVE BAYES CLASSIFIER RESULT BAYES WITH SMOTE

| SMOTE | TP Rate | Precision | Recall | F-Measure |
|---|---|---|---|---|
| 100% | 92.9% | 93.7% | 92.9% | 92.7% |
| 200% | 96.2% | 96.6% | 96.2% | 96.2% |
| 300% | 97.30% | 97.0% | 98.0% | 98.0% |

TABLE V. PART CLASSIFIER RESULT WITH SMOTE

| SMOTE | TP Rate | Precision | Recall | F-Measure |
|---|---|---|---|---|
| 100% | 96.4% | 96.6% | 96.4% | 96.4% |
| 200% | 98.1% | 98.2% | 98.1% | 98.1% |
| 300% | 97.0% | 97.2% | 97.0% | 96.5% |

## V. DISCUSSION

As stated in the results section, the results were obtained by applying SMOTE resampling technique that generates synthetic data samples. Before applying SMOTE, the obtained results were not satisfied for all evaluation measurements. As represented in table III, IV, and V in the results section, significant improvements were seen in all measurement after applying SMOTE.

Table II demonstrates the effectiveness of applying SMOTE at different percentages in improving the performance of FR and PART classifiers, achieving accuracy levels of 99% and 96.7%, respectively. Prior to applying SMOTE, the NB classifier's performance was poor. However, after applying SMOTE at each step, its performance steadily improved. Beyond 300%, however, the accuracy began to decrease due to the generation of an excessive number of synthetic samples for the majority class.

The effectiveness of SMOTE was also evident in other evaluation metrics, including TP Rate, precision, recall, and F-measure. As illustrated in Table III, the RF classifier achieved TP rates of 96.4%, 98.1%, and 99.0% with SMOTE at 100%, 200%, and 300%, respectively. Similarly, the precision values for SMOTE at 100%, 200%, and 300% were 96.6%, 98.2%, and 99.0%, respectively. The recall values for SMOTE at 100%, 200%, and 300% were also high, with the RF classifier achieving 96.4%, 98.1%, and 99.0%, respectively. The F-measure values for SMOTE at 100%, 200%, and 300% were 96.4%, 98.1%, and 99.0%, respectively. These results demonstrate that the RF classifier consistently performed well across all evaluation metrics, indicating the effectiveness of SMOTE in improving its performance.

Table IV presents the results of applying SMOTE at 100%, 200%, and 300% to the NB classifier. The TP rate values for these experiments were 92.9%, 96.2%, and 97.3%, respectively. Similarly, the precision values were 93.7%, 96.6%, and 97.0%, respectively. However, these values were lower than the corresponding precision values for FR and PART classifiers. Likewise, in SMOTE 100% and 200%, the recall and F-measure values for NB were lower than those for FR and PART classifiers.

Table V shows the results for the PART algorithm. In SMOTE 100%, 200%, and 300% the TP rates were 96.4%, 98.1%, and 97%, respectively. The precision values for SMOTE at 100%, 200%, and 300% were also higher than NB, with the PART classifier achieving 96.6%, 98.2%, and 97.2%, respectively. Moreover, the recall values for these experiments were 96.4%, 98.1%, and 97%, respectively. In SMOTE 100%, and 200% the F-Measure values were 96.4% and 98.1%. But in SMOTE 300% it was 96.5%, and it was lower than those for FR and NB classifiers.

In this study, when FR, NB, and PART algorithms were analyzed based on accuracy and other mentioned important measures, and the best result was achieved with FR algorithm for detection of 1000 AFN banknotes. While our model demonstrated good accuracy, it has limitations and not generalized. Specifically, we focused solely on new banknotes, and our dataset included only 1000 AFN notes issued in 1391, without considering other denominations. Due to the scarcity of counterfeit banknotes, it was challenging to create a balanced and representative dataset for our study. Nevertheless, we made every effort to accurately represent the available data, acknowledging the need for further research to expand the scope and generalize the findings of our model.

## VI. CONCLUSION

In this study, we aimed to develop a cost-effective machine learning model to detect counterfeit Afghan banknotes. To achieve this, we utilized image processing techniques to extract key texture features from images of new 1000 AFN banknotes issued in 1391, and built models for FR, Naïve Bayes, and PART classifiers using the WEKA machine learning tool. Our models successfully addressed the issue of imbalanced datasets, with the FR classifier producing the most accurate results.

However, we acknowledge that our model has certain limitations. For instance, it focused solely on detecting counterfeit new 1000 AFN banknotes and may not be generalizable to other denominations. Additionally, the rarity of fake banknotes made it challenging to create a balanced dataset. Despite these challenges, we made every effort to accurately represent the available data. Future work will involve creating a comprehensive dataset from different versions of Afghan banknotes, developing more robust models, and applying the proposed method in real-world applications. We are optimistic that our study represents an important step towards developing a cost-effective and reliable solution for detecting counterfeit Afghan banknotes. To implement our work in a real application in the future, we plan to convert it to a single platform. We will explore two options: translating the MATLAB code to Python code using interpretation techniques [22], or building our models in MATLAB and then deploying them.